\documentclass{bmvc2k}

\title{SOS! : A Streamlined Object-Conditional Transformer for Model-free Segmentation}

\addauthor{Jiaqi Hu}{jiaqi.hu@tum.de}{1,2}
\addauthor{Junwen Huang}{junwen.huang@tum.de}{1,3}
\addauthor{Hongli Xu}{hongli.xu@tum.de}{1}
\addauthor{Peter KT Yu}{peterkty@gmail.com}{4}
\addauthor{Nassir Navab}{nassir.navab@tum.de}{1,3}
\addauthor{Benjamin Busam}{b.busam@tum.de}{1,3}
\addauthor{Slobodan Ilic}{slobodan.ilic@tum.de}{1,2}

\addinstitution{
Technical University of Munich \\
Munich, Germany
}
\addinstitution{
Siemens AG \\
Munich, Germany
}
\addinstitution{
Munich Center for Machine Learning \\
Munich, Germany
}
\addinstitution{
ROBOX \\
Cambridge, MA, USA
}

\runninghead{Hu et al.}{SOS: Single-Reference Object Segmentation}

\usepackage{graphicx}
\usepackage{booktabs}
\usepackage{multirow}
\usepackage{subfiles}
\usepackage{makecell}
\usepackage[accsupp]{axessibility}  
\usepackage{pifont}
\usepackage{hyperref}
\usepackage{orcidlink}

\newcommand{\modelname}{SOS\xspace}

\begin{document}

\maketitle

\begin{abstract}
Foundation segmentation models excel at generating high-quality, class-agnostic masks, but they struggle to associate these proposals with specific target objects. This semantic gap severely hinders their deployment in downstream applications like robotic manipulation, which demand precise unseen objects segmentation. Existing approaches attempt to resolve this by relying on exhaustive 3D object model priors, inherently introducing prohibitive computational overhead and complex, multi-stage pipelines. To address these limitations, we propose \textbf{\modelname} (\textbf{S}treamlined \textbf{O}bject-conditional Transformer for model-free \textbf{S}egmentation). \textbf{\modelname} completely eliminates the reliance on 3D models, requiring only a single reference image per target object. Central to our framework is a novel \textit{Object-Conditional Transformer} that learns \textit{identity-anchored queries}, unifying mask generation and target identification into a single feed-forward pass. This streamlined design drastically improves both structural and computational efficiency. Extensive evaluations across multiple benchmarks demonstrate that \modelname establishes a new state-of-the-art for model-free unseen objects segmentation, delivering accurate and high-efficiency performance. The project page and code are available at \url{https://sos-seg.github.io/}.
\end{abstract}  

\section{Introduction}
Foundation models for segmentation, like SAM family\cite{lin2024sam,ravi2024sam,carion2025sam, zhao2023fast} can segment any object in images. Even though, when coupled with text prompts, they can segment certain object classes, in general settings, they output a large number of segments. Identifying exact segments that belong to the particular object instance is not trivial. This is exactly the first step in tasks like 6D object pose estimation ~\cite{nguyen2024gigapose, labbe2022megapose, ornek2024foundpose, wen2024foundationpose, huang2025raypose, huang2024matchu} and robotic manipulation~\cite{mousavian20196, fang2020graspnet, xu2025funcanon, jiang2023vima}.
Recent research in 6D object pose estimation aims to develop foundation models that can segment and estimate the 6D pose of any object. Such networks are trained once on a large number of objects, each represented by its 3D model, and should generalize to novel objects unseen during training. To know which object to segment and estimate its pose, the majority of methods~\cite{lin2024sam, wen2024foundationpose} accept a CAD model of the unseen object at input together with the query image containing the object that needs to be found. However, in real-world environments, a practical perception system should be able to segment and localize novel targets with minimal prior knowledge. This remains challenging due to the heavy reliance of current methods on extensive object priors, such as CAD models, and complex, multi-stage pipelines.

\begin{figure}[!t]
    \centering
    \includegraphics[width=1.0\textwidth]{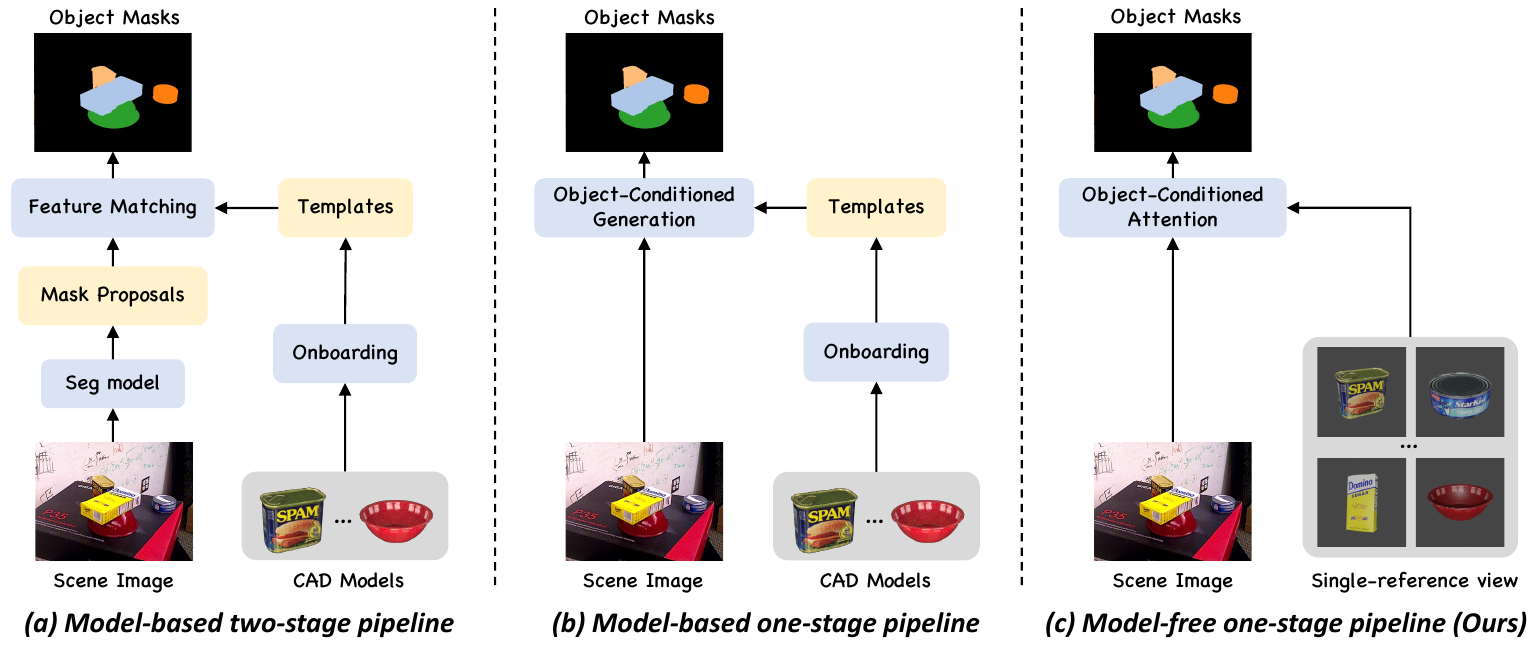}
    \caption{\textbf{Comparison of unseen objects segmentation pipelines.} 
    \textbf{(a) Model-based two-stage pipeline} (e.g., CNOS~\cite{nguyen2023cnos}, SAM6D~\cite{lin2024sam}) uses SAM~\cite{kirillov2023segment} for mask proposals and DINO~\cite{oquab2023dinov2} for feature matching, heavily relying on cumbersome 3D CAD onboarding. 
    \textbf{(b) Model-based one-stage pipeline} (e.g., OC-DiT~\cite{ulmer2025conditional}) predicts masks directly but still requires exhaustive multi-view rendering from 3D CAD priors, while iterative denoising severely bottlenecks inference. 
    \textbf{(c) Model-free one-stage pipeline} (\modelname) completely eliminates 3D CAD reliance. Driven by identity-anchored queries, it requires only a single 2D reference view per object to output identity-aware masks in a single feed-forward pass, enabling real-world deployment.}
    \label{fig:teaser}
\end{figure}

Existing unseen objects segmentation approaches are model-based~\cite{nguyen2023cnos, lin2024sam, cho2025muse, lu2025adapting} and typically employ complex, disjointed two-stage pipelines as shown in Fig.~\ref{fig:teaser}(a). In the first stage, a pre-trained segmentation models ~\cite{kirillov2023segment, zhao2023fast, he2017mask, jocher2023yolo} are used to extract class-agnostic masks. In the second stage, these methods match the generated mask proposals against object templates rendered from multi-view CAD models. This matching process relies on extracting and comparing separate visual descriptors, typically using DINO~\cite{oquab2023dinov2, simeoni2025dinov3}, for both the segments and the templates. This separation fundamentally breaks semantic coherence, since these methods are forced to rely on a cumbersome onboarding stage relying on expensive renderings of 3D CAD models to produce a large number of object templates and establish matches with SAM over-segmentation. 

One-stage approaches like OC-DiT~\cite{ulmer2025conditional} attempt to overcome these disjointed pipelines by predicting masks directly. By unifying the architecture, they successfully preserve semantic coherence and eliminate the need for heuristic matching against external SAM proposals. However, as shown in Fig.~\ref{fig:teaser}(b), OC-DiT still requires excessive rendering of multi-view templates. While this exhaustive 3D prior can compensate for their representational limitations in controlled settings, precise CAD models are inherently difficult to obtain in the wild. Furthermore, its diffusion-based~\cite{peebles2023scalable} architecture requires iterative denoising steps, which significantly increases inference latency.

Overall, whether hindered by disjointed two-stage heuristic matching or slow diffusion mechanisms, the shared reliance on cumbersome CAD-based onboarding introduces significant computational overhead. This precludes fast inference and severely restricts on-the-fly deployment. Consequently, if one attempts to adapt these methods to a practical "model-free" setting by providing only a single reference view, their performance degrades drastically. Stripped of exhaustive 3D information, the inherent representational fragility shared by both paradigms becomes evident. They inherently require CAD-based templates to compensate for their weak feature spaces. In the absence of this multi-view prior, both struggle severely to associate target identities in complex scenes.

To address these issues, we present \modelname, a streamlined, model-free framework for unseen objects segmentation. Instead of relying on extensive 3D priors, \modelname requires only a single RGB scene image and one reference view per target, as depicted in Fig.~\ref{fig:teaser}(c). By treating these reference views as explicit visual conditions, our method establishes an object-conditional Transformer framework driven by \textit{identity-anchored queries}. Consequently, rather than generating class-agnostic proposals via external models for subsequent heuristic matching, \modelname directly outputs \textit{identity-aware} masks in a single feed-forward pass. This unified design simultaneously eliminates the disjointed feature extraction of two-stage pipelines~\cite{nguyen2023cnos, lin2024sam, cho2025muse} and the slow, iterative denoising of single-stage diffusion architectures~\cite{ulmer2025conditional}, yielding a lightweight solution that achieves high accuracy at fast speed.

We evaluate \modelname across multiple benchmarks, where it establishes a new state of the art for model-free segmentation of unseen objects. Notably, \modelname substantially outperforms existing model-based methods when restricted to a single reference image and performs on par with their original setups that rely on complete 3D CAD models.

In summary, our main contributions are: 
\begin{enumerate}
    \item \textbf{Unified Model-Free Framework:} We propose \modelname, a streamlined, one-stage architecture for unseen objects segmentation. Conditioned only on a single reference view per object, \modelname completely eliminates both object CAD onboarding phase and the reliance on external region proposal networks.
    
    \item \textbf{Object-Conditional Transformer:} We introduce a novel architecture to learn the \textit{identity-anchored query} that unifies mask generation and instance identification into a single feed-forward pass. This design concurrently localizes and identifies the objects in the scene, achieving high-efficient inference while bypassing the bottlenecks of iterative denoising or dense template matching.
    
    \item \textbf{Computational Efficiency and Robustness.} We present an efficient network with minimal inference latency, achieving competitive performance on multiple benchmarks compare to with state-of-the-art model-based baselines, and demonstrates robustness to clutter in "in-the-wild" scenarios.
    
\end{enumerate}

    
    

\label{sec:intro}

\section{Related Work}

\subsection{Model-Based segmentation of unseen objects}
The task of unseen objects segmentation requires a perception model to localize and segment novel targets without prior object-specific training. To identify these unknown targets, the perception model must rely on explicit object priors. Driven by this requirement, most existing methods~\cite{nguyen2023cnos, lin2024sam, ulmer2025conditional, lu2025adapting} adopt a model-based paradigm, demanding full 3D CAD models of the targets to render comprehensive offline reference templates.

To utilize these templates during inference, CNOS~\cite{nguyen2023cnos} introduced what has become the standard two-stage pipeline. The approach is straightforward: a foundation model like SAM~\cite{kirillov2023segment} extracts class-agnostic masks, and a separate descriptor network like DINOv2~\cite{oquab2023dinov2} subsequently matches them against the CAD templates to assign identities. Because this hard decoupling of localization and identification is inherently prone to correspondence errors, later works have focused heavily on patching the matching step. SAM-6D~\cite{lin2024sam}, for example, attempts to refine the matching stage of CNOS with novel appearance and geometric matching scores. MUSE~\cite{cho2025muse} takes this a step further, deploying GroundingDINO~\cite{liu2024grounding} to filter out background clutter before routing the remaining masks through an even denser matching mechanism. Ultimately, compensating for a disjointed architecture by stacking heavier models and convoluted rules creates severe computational bottlenecks, making real-world deployment hard.

In an attempt to bypass these fragmented pipelines, OC-DiT~\cite{ulmer2025conditional} recasts the segmentation problem as a conditional diffusion process. While this generative approach successfully eliminates explicit mask matching, it cannot escape the fundamental reliance on dense 3D CAD priors. More critically, the iterative denoising required by diffusion introduces massive inference delays, effectively trading one computational bottleneck for another.

 \textbf{\modelname} addresses these limitations by shifting to a purely transformer-based~\cite{vaswani2017attention}, one-stage architecture. Unlike previous methods, \modelname operates without 3D CAD models, relying instead on a single reference view to condition the segmentation process. This design eliminates the primary sources of inference latency and correspondence errors, resulting in a simple, real-world solution suitable for practical deployment.

\subsection{Visual Reference Segmentation}
Visual reference segmentation has emerged as a powerful paradigm for isolating specific objects guided by visual prompts. While rooted in Referring Image Segmentation (RIS), which traditionally centers on language-to-image grounding, this line of research focuses on using images as query prompts to define the target. Current state-of-the-art methods, such as \cite{hossain2024visual, suo2024rethinking, zhang2024bridge}, predominantly emphasize category-level generalization \cite{hong2022cost, peng2023hierarchical, min2021hypercorrelation, liu2023matcher, wang2023seggpt}. By treating the reference image as a semantic prototype, these approaches successfully identify all objects within a broader class. However, this focus on category-level consistency often comes at the expense of instance-level identity, leading models to ignore distinctive visual cues such as unique textures, brand logos, or subtle wear-and-tear patterns.

This limitation is particularly evident in unseen objects segmentation, a task that inherently demands instance-specific identification. Unlike category-oriented frameworks that might treat any cup as a valid match, \modelname leverages the visual prompt as a unique identity constraint to isolate the exact individual shown. By shifting the objective from "finding any object of a class" to "distinguishing a specific target from intra-class distractors," our approach achieves a level of fine-grained discrimination previously overlooked. This capability is essential for practical deployment in robotic manipulation and long-term object tracking, where interacting with a specific entity is far more critical than recognizing a general category.

\label{sec:related_work}

\section{Methodology}
\modelname is a model-free framework designed for unseen objects segmentation using only one reference view per object as condition. We introduce a novel \textbf{Object-conditional Transformer} to learn the \textit{Identity-Anchored Query} that unifies the mask generation and target identification into a single feed-forward pass. An overview of our pipeline is illustrated in Fig.~\ref{fig:pipeline}.

\subsection{Problem Formulation}
Our framework aims to achieve one-shot semantic segmentation of arbitrary objects within a scene, guided by a single reference exemplar, while maintaining the ability to process multiple objects within a single forward pass. Formally, given a target RGB scene $I_s \in \mathbb{R}^{H \times W \times 3}$ and a set of $N$ object conditions $\mathcal{C} = \{(C_i, M_{c,i})\}_{i=1}^N$, where each $C_i \in \mathbb{R}^{H_i \times W_i \times 3}$ denotes a reference image and $M_{c,i} \in \{0, 1\}^{H_i \times W_i}$ defines its corresponding foreground mask, the model predict a set of binary masks $\mathcal{M} = \{\mathbf{M}_i\}_{i=1}^N$. Each predicted mask $\mathbf{M}_i \in \{0, 1\}^{H \times W}$ represents the 2D pixel occupancy of the $i$-th target object within $I_s$. To ensure robustness against false positives, the model explicitly handles \textit{negative conditions}: if the target object is absent from the scene, the framework is expected to yield a null response, resulting in an empty mask $\mathbf{M}_i = \mathbf{0}$.

\subsection {Identity-Anchored Query}
Human visual search is inherently target-driven; rather than exhaustively parsing every entity in a cluttered scene, the human visual system leverages a prototypical representation of a target to selectively modulate the visual field. Inspired by this cognitive efficiency, we rethink the standard query-based formulation in DETR~\cite{carion2020end}. Current DETR-like architectures~\cite{dai2021dynamic, zhang2022dino, meng2021conditional, zhu2020deformable, cheng2022masked} typically employ \textit{anonymous queries} constrained by learned spatial priors. This "segment-then-classify" paradigm necessitates a decoupled process: the model first generates class-agnostic mask proposals and subsequently relies on a classification head for post hoc identity assignment. While effective for closed-set semantic segmentation where inter-category variance is high, this approach is sub-optimal for one-shot tasks. Unseen objects segmentation requires identifying the \textit{exact same instance}, demanding precise intra-category discrimination against visually similar distractors. In current architectures, the visual reference is often injected late in the pipeline, depriving the mask generation process of early-stage semantic guidance. Consequently, anonymous queries lack the discriminative signal required for fine-grained instance matching in unconstrained scenes.

To address this bottleneck, we propose the \textit{Identity-Anchored Query}, a novel representation that shifts the paradigm from class-agnostic proposal generation to target-specific referencing through the proposed Object-conditional Transformer. The network learns a universal semantic probe that undergoes dual path updates: it absorbs the reference identity while simultaneously grounding the target within the scene. This unified reasoning merges mask generation and identity verification into a single step. In the end, the probe becomes a deterministic descriptor that predicts masks directly, removing the need for post hoc classification. We will discuss the technical details in Sec~\ref{sec:object-condition transformer}

\subsection{Network Architecture}
Our framework for unseen objects segmentation consists of three core components:
(1) a \textbf{Feature Encoder} that extracts scene and condition tokens;
(2) an \textbf{Object-conditional Transformer} that progressively transforms the universal semantic cue into an \textit{Identity-Anchored Query} by concurrently injecting target reference features and retrieving scene context; and
(3) a \textbf{Mask Decoder} that decodes these identity-aware descriptors into precise, instance-aware segmentation masks.

\begin{figure}[!t]
    \centering
    \includegraphics[width=1.0\textwidth]{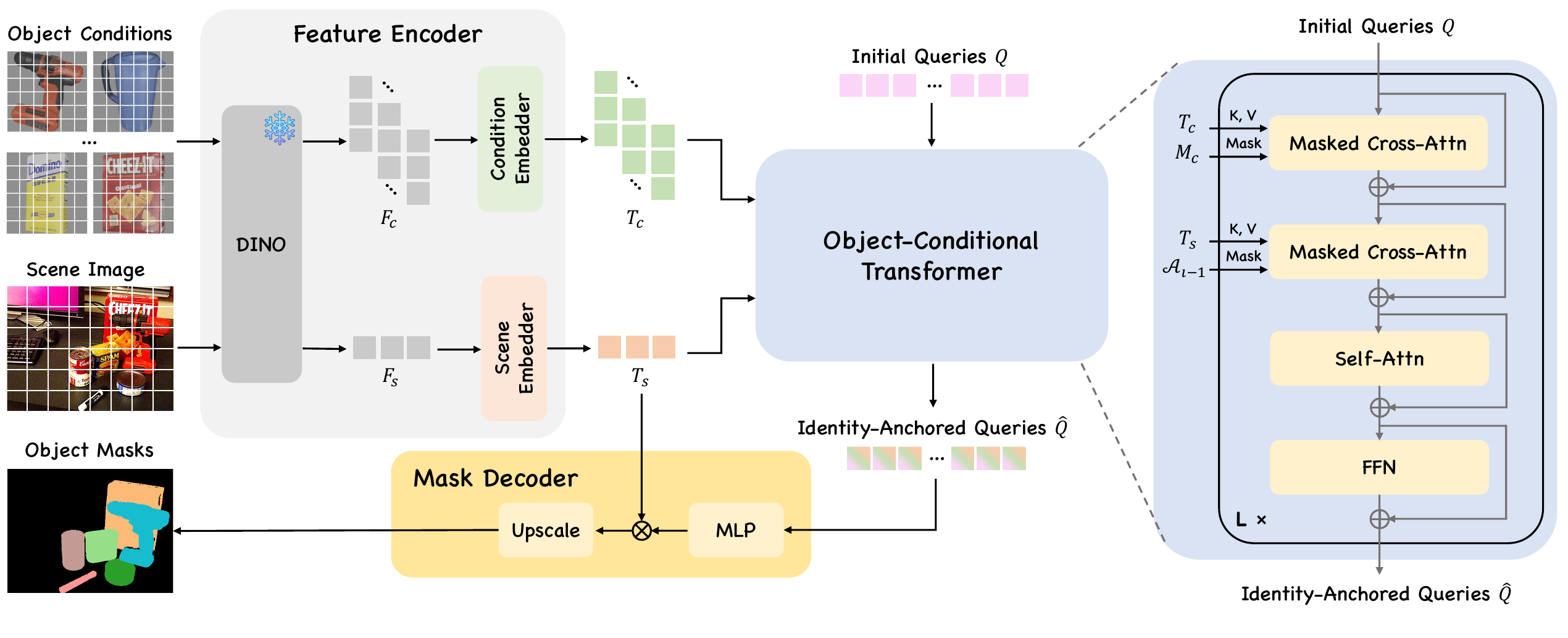}
    \caption{\textbf{Overview of the \modelname framework.} 
    (1) \textbf{Feature Encoder}: A frozen DINO backbone extracts feature maps ($F_s, F_c$) from the scene and object conditions respectively. The Scene and Condition Embedders subsequently process these maps into token sequences ($T_s, T_c$) augmented with spatial and identity positional encodings. 
    (2) \textbf{Object Conditional Transformer}: Initial queries $Q$ iteratively evolve into deterministic identity-anchored queries. Within each block, queries perform masked cross attention with the condition tokens $T_c$ and condition masks $M_c$ to absorb target identities, followed by masked cross attention with the scene tokens $T_s$ restricted by the preceding attention masks $\mathcal{A}_{l-1}$ for spatial grounding, and self attention to resolve overlapping targets and occlusions.
    (3) \textbf{Mask Decoder}: The identity-anchored queries $\hat{Q}$ are decoded, interacting with the scene tokens to directly output the final masks for each object.}
    \label{fig:pipeline}
\end{figure}

\subsubsection{Feature Encoder.} 
Given a scene image $I_s \in \mathbb{R}^{H \times W \times 3}$ and a set of $N$ object conditions $\{C_i\}_{i=1}^N$, we first project them into a shared semantic space using a pre-trained ViT~\cite{dosovitskiy2020image} encoder which partitions the scene and condition images into non-overlapping patches to extract dense feature maps. Let $F_s \in \mathbb{R}^{h \times w \times d}$ denote the scene feature map and $F_c \in \mathbb{R}^{N \times h' \times w' \times d}$ denote the stacked object condition feature maps, where $(h, w)$ and $(h', w')$ represent the respective feature map resolutions and $d$ is the embedding dimension. 

\textbf{\textit{Scene Embedder.}} The scene feature map $F_s \in \mathbb{R}^{h \times w \times d}$ is flattened into scene tokens $T_s \in \mathbb{R}^{S \times d}$. To preserve the structural layout, we augment these tokens by directly adding a learnable positional embedding $P_s \in \mathbb{R}^{1 \times S \times d}$. This explicit coordinate injection provides the spatial anchoring necessary for precise object localization.

\textbf{\textit{Condition Embedder.}} Similarly, the condition feature map $F_c \in \mathbb{R}^{N \times h' \times w' \times d}$ is flattened into condition tokens $T_c \in \mathbb{R}^{N \times S' \times d}$, where $S' = h' \times w'$ and $N$ denotes the distinct object conditions. To uniquely identify each target, we augment these tokens with two learnable positional embedding tensors: a spatial embedding $P_{c,patch} \in \mathbb{R}^{1 \times 1 \times S' \times d}$ preserving the internal layout of each reference view, and an identity embedding $P_{c,id} \in \mathbb{R}^{1 \times N \times 1 \times d}$ differentiating the $N$ conditions. The final condition tokens are obtained via element wise addition utilizing tensor broadcasting.

\subsubsection{Object-conditional Transformer.}
\label{sec:object-condition transformer}
We first define a learnable universal semantic probe $q \in \mathbb{R}^{1 \times d}$. By replicating this probe $N$ times, we generate the initial queries $Q \in \mathbb{R}^{N \times d}$ to match the number of object conditions. As illustrated in Fig.~\ref{fig:pipeline}, the Object Conditional Transformer transforms these initial queries into deterministic identity-anchored queries $\hat{Q}$. Through continuous dual path fusion with both condition and scene tokens, the network gradually implants target identities into the queries while learning their spatial boundaries. This process outputs the final identity-anchored queries $\hat{Q}$ for mask decoding.

Within each block $l \in \{1, 2, \dots, L\}$, the intermediate queries $\hat{Q}^{(l-1)}$ (where $\hat{Q}^{(0)} = Q$) perform masked cross attention with their assigned condition tokens $T_{c,i}$ and condition masks $M_{c,i}$. This operation explicitly extracts target identities while preventing crosstalk between conditions. To efficiently localize targets and suppress distractors, these identity enriched queries then attend to the scene tokens $T_s$ via masked cross attention. This grounding step is restricted by a boolean mask $\mathcal{A}^{(l-1)}$ derived from the preceding block, with the initial block attending globally. The spatially grounded queries subsequently exchange contextual information via self attention to resolve overlapping targets and occlusions. A Feed Forward Network (FFN) then processes these features to yield the refined identity-anchored queries $\hat{Q}^{(l)}$.

\subsubsection{Mask Decoder.}
After each transformer block $l$, we decode the refined identity-anchored queries $\hat{Q}^{(l)}$ to compute the attention mask $\mathcal{A}^{(l)}$ for the next block. We pass the queries through a lightweight MLP and compute the spatial dot product with the scene tokens $T_s$:$$P^{(l)} = \text{MLP}(\hat{Q}^{(l)}) \otimes T_s$$where $\otimes$ denotes the spatial dot product producing the mask logits $P^{(l)}$. Following standard masked attention~\cite{cheng2022masked}, we binarize these logits via sigmoid activation and thresholding to obtain the mask prediction $\mathbf{M}^{(l)}$. The boolean attention mask $\mathcal{A}^{(l)}$ at spatial location $(x, y)$ is defined as:$$\mathcal{A}^{(l)}(x, y) = \begin{cases} 0 & \text{if } \mathbf{M}^{(l)}(x, y) = 1 \\ -\infty & \text{otherwise} \end{cases}$$ Adding this mask to the attention logits before the softmax operation restricts the cross attention span entirely to the predicted foreground regions. 

After the final block $L$, we upscale the mask logits $P^{(L)}$ to the target image dimensions. These high resolution logits are then passed through a sigmoid function and thresholded to output the final mask predictions $\mathcal{M} = \{\mathbf{M}_i\}_{i=1}^N \in \{0, 1\}^{N \times H \times W}$.

\subsection{Training Objectives}
Our \textit{Identity-Anchored Query} formulation eliminates the bipartite matching (e.g., the Hungarian algorithm) required by conventional DETR~\cite{dai2021dynamic, zhang2022dino, zhu2020deformable} architectures. Because each query is deterministically anchored to an input condition, the predicted masks naturally maintain a one to one correspondence with the ground truth.
With this direct correspondence established, we drop the classification loss entirely and rely solely on mask supervision. We adopt the mask loss from Mask2Former~\cite{cheng2022masked}, which combines Binary Cross Entropy (BCE) and Dice loss~\cite{milletari2016v}. The mask loss for block $l$ is:$$\mathcal{L}^{(l)} = \lambda_{\text{bce}} \mathcal{L}_{\text{bce}}^{(l)} + \lambda_{\text{dice}} \mathcal{L}_{\text{dice}}^{(l)}$$where $\lambda_{\text{bce}}$ and $\lambda_{\text{dice}}$ denote the loss weights.

To stabilize training and supervise the intermediate boolean attention masks, we employ deep supervision across our transformer architecture by defining the total objective as: $$\mathcal{L}_{\text{total}} = \sum_{l \in \{2, 4, 6, 8\}} \mathcal{L}^{(l)}$$where $l$ represents the one-based index of the supervised block.

\label{sec:3_method}

\section{Experiments}
\label{sec:experiment}

\subsection{Datasets and Evaluation metric.} We evaluate our method on four core BOP challenge datasets~\cite{sundermeyer2023bop}: YCB-Video (YCB-V)~\cite{xiang2017posecnn}, TUD-L~\cite{hodan2018bop}, LineMod Occlusion (LM-O)~\cite{brachmann2014learning}, and HomebrewedDB (HB)~\cite{kaskman2019homebreweddb}. Combined, these benchmarks contain 65 objects in cluttered and occluded scenes. The object set covers a wide range of properties, including textured or untextured surfaces, symmetric or asymmetric geometries, and both household and industrial categories.

We evaluate our method using the Average Precision (AP) metrics, following the COCO metric and the BOP challenge evaluation protocol\cite{sundermeyer2023bop}. The AP metric is calculated as the mean of AP values at different Intersection over Union (IoU) thresholds ranging from 0.50 to 0.95 with an increment of 0.05.

\subsection{Implementation Details}
\subsubsection{Architecture.} \modelname is built upon a pre-trained, frozen DINOv3-L/16~\cite{simeoni2025dinov3} backbone. To balance computational efficiency with feature granularity, scene images are processed at $480 \times 640$ resolution, while condition images are resized to $224 \times 224$. These features are integrated via an object-conditioned transformer consisting of 8 blocks with a hidden size of 1024 and 16 attention heads. A 2-layer MLP is appended as the mask decoder to produce the final segmentation.

\subsubsection{Training data.} We train the model using large synthetic dataset provided by FoundationPose~\cite{wen2024foundationpose}. The training assets are sourced from Objaverse~\cite{deitke2023objaverse} and GSO~\cite{downs2022google}, comprising approximately 41K unique objects in total. The full dataset consists of roughly 680K scenes, each containing multiple objects in cluttered environments.

\subsubsection{Data Augmentation.} To improve generalization, we apply both geometric and photometric augmentations. For geometric transforms, images are randomly cropped with a probability of 0.8 at a scale between $[0.7, 0.99]$, followed by random horizontal flips. Photometric augmentations include random adjustments to brightness, contrast, sharpness, and color, alongside Gaussian blur, to enhance the model's robustness to varying lighting and imaging conditions.

\subsubsection{Training Setup.} 
To enable the model to handle an arbitrary number of conditions, we employ a dynamic condition sampling strategy. For each training sample, we randomly select $n \in [1, 20]$ object conditions, including both positive and negative ones. To enhance robustness against viewpoint variations, each condition is instantiated by drawing a single image from a predefined set of 12 rendered perspectives.

The entire training process spans 120K steps with a total batch size of 16 using the AdamW optimizer~\cite{loshchilov2017decoupled}. We set the initial learning rate to $1 \times 10^{-4}$, beginning with a 2K-step linear warmup. This learning rate is maintained constantly until 70K steps, after which a cosine annealing schedule is applied to gradually decay it to a final value of $1 \times 10^{-6}$. Training is conducted on 2 NVIDIA RTX 4090 GPUs and takes approximately 25 hours.

\subsection{Main Results}

\begin{table}[!htbp]
\centering
\resizebox{\textwidth}{!}{
\begin{tabular}{l | c | c | c | ccccc | c}
\toprule
Method & Model-based & \begin{tabular}{@{}c@{}}Onboarding \\ (s / per-obj)\end{tabular} & \begin{tabular}{@{}c@{}}Ref. \\ Image(s)\end{tabular} & YCB-V & TUD-L & LM-O & HB & Average & \begin{tabular}{@{}c@{}}Inference \\ (s / per-img)\end{tabular} \\ \midrule
$^*$CNOS~\cite{nguyen2023cnos}    & \checkmark & 18.82 & 42 - CAD & 59.9 & 48.0 & 39.7 & 51.1 & 49.7 & \underline{0.221} \\
$^*$SAM6D~\cite{lin2024sam}       & \checkmark & 18.82 & 42 - CAD & 59.5 & 49.8 & 44.4 & 55.7 & 52.3 & 0.249 \\
$^*$NIDS-Net\cite{lu2025adapting} & \checkmark & 18.82 & 42 - CAD & 65.0 & 55.6 & 43.9 & 62.0 & 56.6 & 0.485 \\
$^*$MUSE\cite{cho2025muse}        & \checkmark & 18.82 & 42 - CAD & \underline{69.0} & 57.3 & \underline{47.7} & \underline{63.5} & \textbf{59.3} & 0.505 \\
$^*$OC-DiT~\cite{ulmer2025conditional} & \checkmark & 18.82 & 42 - CAD & \textbf{71.7} & \textbf{59.4} & 40.1 & 61.5 & 58.2 & - \\ \hline

CNOS  & $\times$ & 0.0 & 1 & 55.3 & 45.2 & 41.1 & 42.5 & 46.0 & - \\
SAM6D & $\times$ & 0.0 & 1 & 56.3 & 47.9 & 41.8 & 44.7 & 47.6 & - \\
Ours  & $\times$ & \textbf{0.0} & 1 & 61.7 & \underline{57.6} & \textbf{50.3} & \textbf{65.6} & \underline{58.8} & \textbf{0.120} \\  
\bottomrule
\end{tabular}
}
\vspace{2mm}
\caption{\textbf{Results on the BOP challenge datasets.} We report the AP for the task unseen objects segmentation. The highest value in each column is denoted in \textbf{bold}, and the second highest is \underline{underlined}. For baseline methods requiring CAD models, the reported onboarding time includes both template rendering via BlenderProc~\cite{denninger2019blenderproc} and templates feature extraction, measured on a single NVIDIA RTX 4090 GPU. \textit{AP and inference time of methods marked with $^*$ are sourced directly from the original papers.}}
\label{tab:bop-results}
\end{table}

We evaluate our method in a model free setup against CNOS~\cite{nguyen2023cnos} and SAM6D~\cite{lin2024sam}. Because these baselines typically rely on multiple templates, we adapt them to support single reference view inputs. To ensure a fair comparison, we evaluate all methods using the exact same reference images provided to \modelname, which are directly cropped from real scenes. \textit{Visualizations of all reference objects are provided in the supplementary material.}

\subsubsection{Accuracy.} 
\begin{figure}[!htbp]
    \centering
    \includegraphics[width=1.0\textwidth]{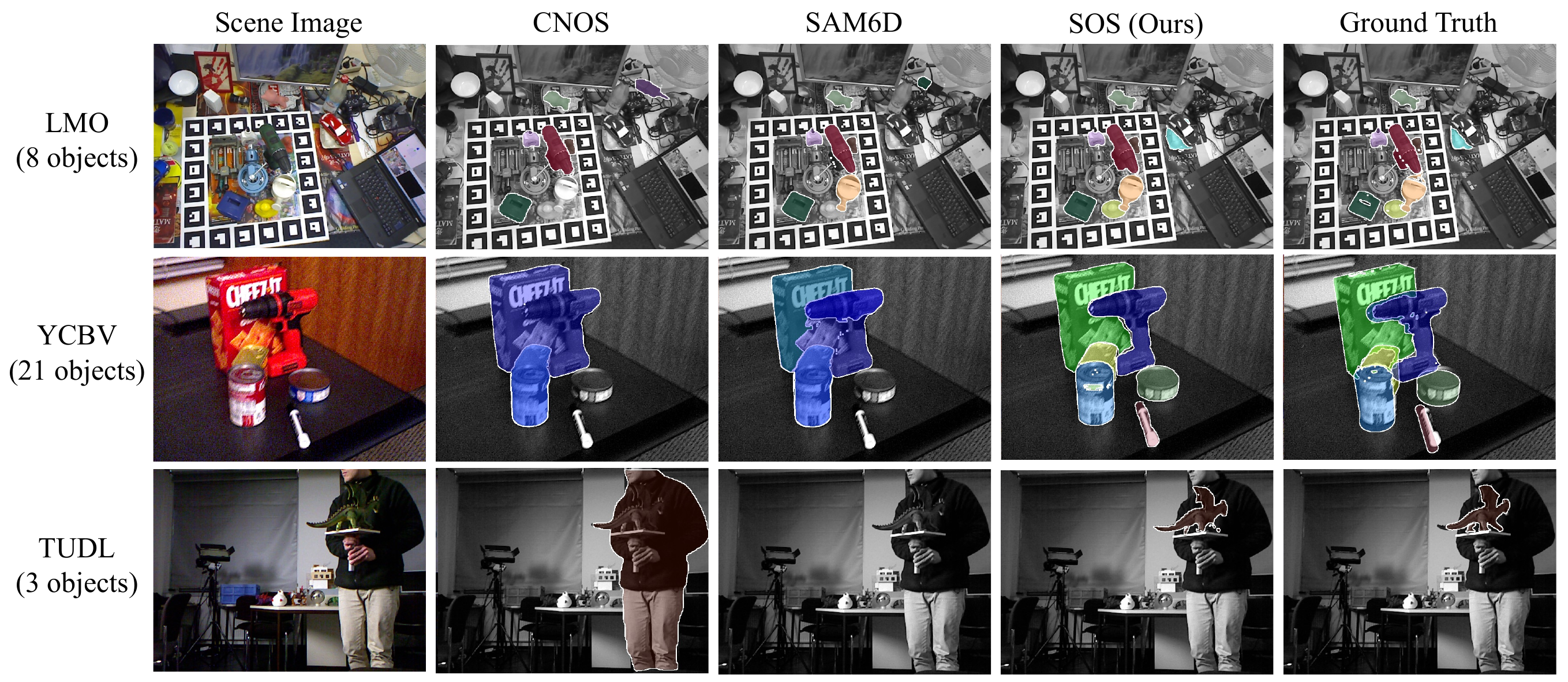}
    \caption{\textbf{Qualitative results on LM-O~\cite{brachmann2014learning}, YCB-V~\cite{xiang2017posecnn} and TUD-L~\cite{hodan2018bop}.} The first column shows the input RGB image and the following columns depict the detections produced by CNOS~\cite{nguyen2023cnos}, SAM6D~\cite{lin2024sam} and \modelname with confidence scores greater than 0.5. The last column shows the Ground Truth.}
    \label{fig:Qualitative results}
\end{figure}

Tab.~\ref{tab:bop-results} reports the quantitative results on YCB-V~\cite{xiang2017posecnn}, TUD-L~\cite{hodan2018bop}, LM-O~\cite{brachmann2014learning}, and HB~\cite{kaskman2019homebreweddb} and Fig.~\ref{fig:Qualitative results} shows the qualitative results. Under the single-view setting, our method significantly outperforms adapted baselines. \modelname achieves an average AP of 58.8 compared to 46.0 for CNOS~\cite{nguyen2023cnos} and 47.6 for SAM6D~\cite{lin2024sam}, yielding an absolute improvement of +11.2 AP over the strongest baseline. 
Furthermore, despite relying on only one reference image, \modelname achieves performance highly competitive with state-of-the-art methods that require 42 CAD-rendered templates. Notably, it sets a new state-of-the-art on the heavily occluded LM-O dataset (50.3 AP) and the HB dataset (65.6 AP), surpassing all model-based baselines. This confirms that our model effectively extracts robust identity priors from a single view, eliminating the necessity for exhaustive templates and CAD models.

\subsubsection{Speed.} 
\modelname achieves an average inference latency of 0.120 seconds per image with multiple target objects, using a single NVIDIA RTX 4090 GPU and outperforming all evaluated baselines. It is nearly twice as fast as CNOS (0.221s) and over four times faster than recent state-of-the-art models like NIDS-Net (0.485s) and MUSE (0.505s). 
More importantly, the reported inference times for model-based methods exclude their offline onboarding stage. As shown in Tab.~\ref{tab:bop-results}, these baselines require an additional 18.82 seconds per object to render CAD models and extract multi-template features. This initialization overhead scales linearly with the number of targets, severely limiting their applicability in dynamic environments. \modelname, being entirely model-free, reduces this preparation cost to zero, ensuring highly efficient real-world deployment.

\subsection{Ablation Study}

\subsubsection{Architecture ablation.}
We conduct ablations on different architectural design choices of SOS across multiple BOP benchmarks, while keeping the training data and training strategy fixed.

First, we investigate how the attention formulations within our decoder affect the overall framework, focusing on cross-attention with scene tokens and self-attention among identity-anchored queries. Our experimental results indicate that both components are indispensable, as removing either one leads to a significant reduction in accuracy. Eliminating cross-attention with scene tokens results in a massive drop of 15.8 points, demonstrating that the network relies heavily on this cross-modal interaction to ground conditional prompts into the scene image. Furthermore, discarding the self-attention mechanism among the queries causes a 6.2 points decrease in performance. This drop highlights the necessity of contextualized reasoning among identity queries, which prevents redundant predictions and stabilizes instance-level identification.

\begin{table}[!htbp]
\centering
\begin{tabular}{l c c c c c}
\toprule
Configuration & YCB-V & TUD-L & LM-O & HB & Average \\
\midrule
w/o self-attn & 59.3 & 50.3 & 44.3 & 56.6 & 52.6 $\downarrow_{6.2}$ \\
w/o scene cross-attn & 41.0 & 51.2 & 31.2 & 48.4 & 43.0 $\downarrow_{15.8}$ \\
\midrule
w/o deep supervision & 47.0 & 52.0 & 31.1 & 48.2 & 44.6 $\downarrow_{14.2}$ \\
\midrule
\makecell[l]{w/ anonymous query \\ (instead of identity-anchored query)} & 47.0 & 25.8 & 37.5 & 51.4 & 40.4 $\downarrow_{18.4}$ \\
\midrule
\textbf{Full model} & \textbf{61.7} & \textbf{57.6} & \textbf{50.3} & \textbf{65.6} & \textbf{58.8} \\
\bottomrule
\end{tabular}
\vspace{2mm}
\caption{\textbf{Ablation study of SOS architectural design choices on BOP benchmarks.}
We report AP for unseen object segmentation on YCB-V, TUD-L, LM-O, and HB, together with the average AP. All ablation variants are trained on the same dataset for the same number of iterations.}
\end{table}

We further examine the effectiveness of the proposed identity-anchored queries by comparing them with anonymous queries. To ensure a fair comparison with minimal architectural changes, we adopt the similar design of Mask2Former~\cite{cheng2022masked} for the anonymous query variant. Specifically, we initialize N=100 learnable queries, remove cross-attention with conditional tokens, and instead employ separate mask and classification heads for mask prediction and classification. Consequently, this modification triggers the most significant performance drop among all ablation studies, with a steep decline of 18.4 (yielding 40.4). This massive gap directly confirms the effectiveness of our identity-anchored queries, demonstrating their substantial performance gain over traditional anonymous queries.

Additionally, we investigate the impact of the auxiliary training strategies. We evaluate the necessity of deep supervision by restricting the loss computation exclusively to the final layer output. This modification leads to a significant decline in accuracy, confirming that supervising intermediate decoder layers is essential for guiding progressive mask refinement and stabilizing the optimization process.

\begin{table}[!htbp]
\centering
\begin{tabular}{l|c|ccccc}
\toprule
Method & Backbone & YCB-V & TUD-L & LM-O & HB & Average \\ 
\midrule
CNOS & DINOv3 & 63.6 & 48.3 & 43.7 & 50.5 & 51.5 \\ 
\hline
SAM6D & DINOv3 & \textbf{63.7} & 43.6 & 43.7 & 50.3 & 50.3 \\ 
\hline
Ours & DINOv3 & 61.7 & \textbf{57.6} & \textbf{50.3} & \textbf{65.6} & \textbf{58.8}  \\ 
\bottomrule
\end{tabular}
\vspace{2mm}
\caption{\textbf{Backbone ablation under the single-reference setting.} CNOS, SAM6D, and SOS are compared with DINOv3 using the same real cropped reference images.}
\label{tab:ablation-backbone}
\end{table}

Finally, we investigate the impact of the backbone. Since CNOS and SAM6D originally use DINOv2~\cite{oquab2023dinov2}, we replace their default backbone with DINOv3~\cite{simeoni2025dinov3} for a fair comparison under the same single-reference setting. As shown in Tab.~\ref{tab:ablation-backbone}, SOS consistently outperforms CNOS and SAM6D on most datasets and achieves the best average performance. This demonstrates that SOS can better benefit from the stronger representation capabilities and richer semantic features of DINOv3, which are essential for precise object-centric matching.

\subsubsection{Efficiency.}

\begin{figure}[!htbp]
    \centering
    \begin{minipage}[c]{0.48\textwidth}
        \centering
        \includegraphics[width=\textwidth]{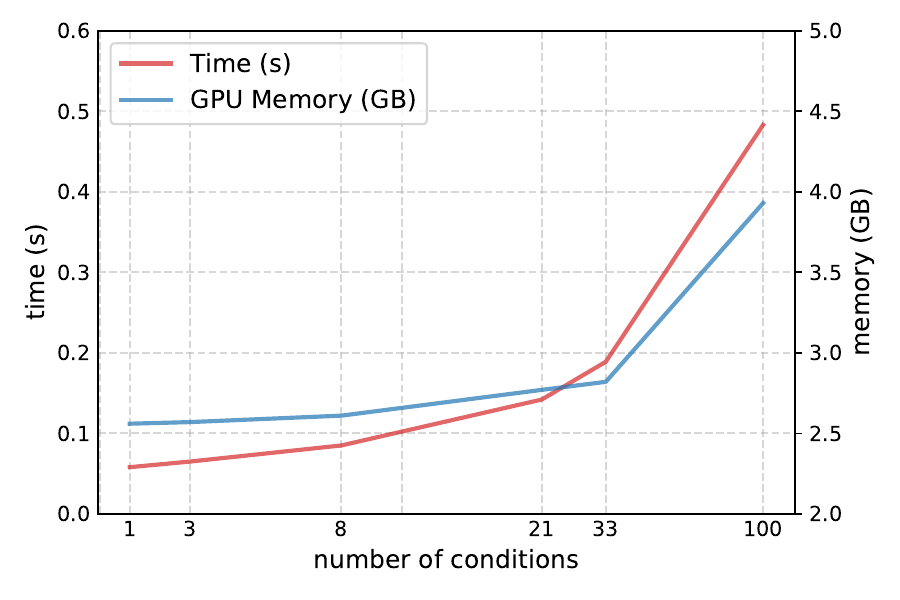}
        \caption{\textbf{Inference Efficiency.} Time and memory consumption during inference across varying numbers of input conditions.}
        \label{fig:efficiency}
    \end{minipage}
    \hfill
    \begin{minipage}[c]{0.48\textwidth}
        \centering
        \includegraphics[width=\textwidth]{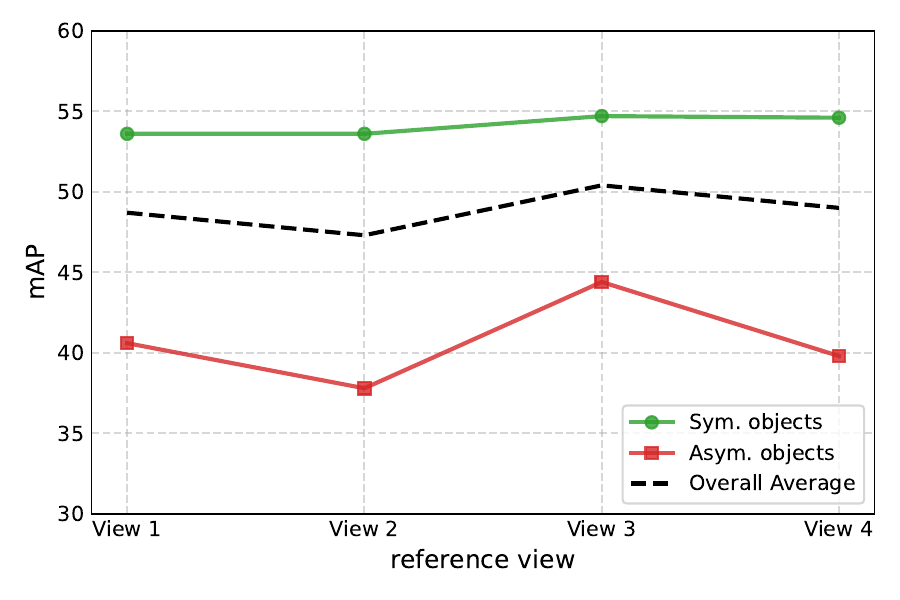}
        \caption{\textbf{Condition sensitivity.} The performance variance for symmetric objects, asymmetric objects and overall average across different reference views.}
        \label{fig:view_sensitive}
    \end{minipage}
\end{figure}

We evaluate the relationship between the number of input conditions and inference efficiency in Fig .~\ref{fig:efficiency}. Our model operates at 20 FPS with a single object condition and maintains 10 FPS when simultaneously processing 10 conditions. Coupled with a consistently low GPU memory footprint, these results demonstrate that our architecture is practical for real world deployment.

\subsubsection{Condition Sensitivity.}
\begin{figure}[!htbp]
    \centering
    \includegraphics[width=1.0\textwidth]{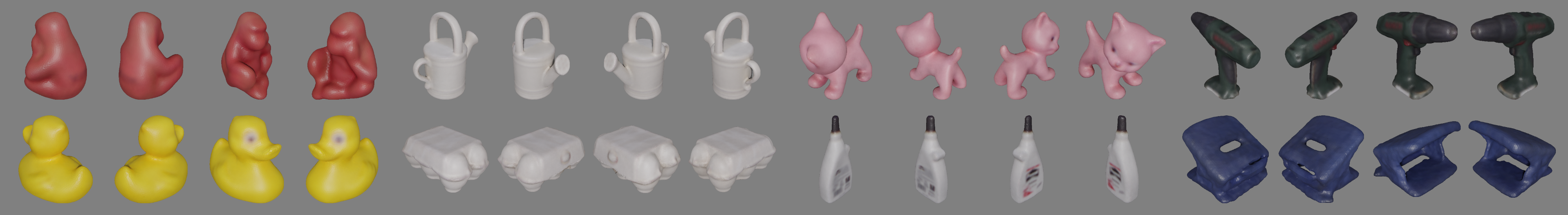}
    \caption{\textbf{Visual condition templates.} This grid illustrates the four distinct rendered perspectives used for our sensitivity analysis on the LM-O~\cite{brachmann2014learning} dataset. For each object, the images arranged from left to right correspond sequentially to view 1, view 2, view 3, and view 4 in Tab.~\ref{tab:ablation_lmo}.}
    \label{fig:ablation_lmo_grid}
\end{figure}

Intuitively, segment unseen objects from a single reference view inherently exposes the model to performance variations based on the chosen perspective. To quantify how these visual conditions influence accuracy, we evaluate our model on the LM-O~\cite{brachmann2014learning} dataset using different reference views. For each object, we render four distinct perspectives at azimuth angles of 45, 135, 225, and 315 degrees (Fig.~\ref{fig:ablation_lmo_grid}). By independently measuring the average precision for each view, we isolate the impact of the visual condition.

Results in Tab.~\ref{tab:ablation_lmo} and Fig.~\ref{fig:view_sensitive} confirm that performance varies across different reference perspectives. This sensitivity is fundamentally tied to object geometry. To analyze this effect, we categorize the objects into symmetric (\textit{can, driller, duck, eggbox, glue}) and asymmetric (\textit{ape, cat, holepuncher}) groups.

\begin{table}[!htbp]
    \centering
    \resizebox{\textwidth}{!}{
        \begin{tabular}{l | ccc | c | ccccc | c | c}
            \toprule
            \multirow{2}{*}{View} & \multicolumn{4}{c|}{Asymmetric Objects} & \multicolumn{6}{c|}{Symmetric Objects} & \multirow{2}{*}{\textbf{Overall Avg.}} \\ \cmidrule(lr){2-5} \cmidrule(lr){6-11}
            & ape & cat & holepuncher & \textbf{Avg.} & can & driller & duck & eggbox & glue & \textbf{Avg.} & \\ \midrule
            view 1 & 39.8 & 47.2 & 34.7 & \textbf{40.6} & 55.8 & 63.8 & 58.9 & 47.6 & 41.7 & \textbf{53.6} & 48.7 \\
            view 2 & 30.8 & 46.5 & 36.2 & \textbf{37.8} & 56.4 & 64.1 & 58.0 & 47.5 & 42.0 & \textbf{53.6} & 47.3 \\
            view 3 & 45.6 & 41.2 & 46.3 & \textbf{44.4} & 56.1 & 64.0 & 62.2 & 48.2 & 43.1 & \textbf{54.7} & 50.4 \\
            view 4 & 45.1 & 44.0 & 30.3 & \textbf{39.8} & 55.9 & 64.3 & 62.0 & 47.5 & 43.1 & \textbf{54.6} & 49.0 \\ \midrule
            $\Delta$ (Max - Min) & 14.8 & 6.0 & 16.0 & \textbf{6.6} & 0.6 & 0.5 & 4.2 & 0.7 & 1.4 & \textbf{1.1} & 3.1 \\
            \bottomrule
        \end{tabular}
}
\vspace{2mm}
    \caption{\textbf{Ablation on view condition sensitivity.} We report the average precision across different azimuth angles for eight objects from the LM-O~\cite{brachmann2014learning} dataset. Symmetric objects maintain stable scores, while asymmetric objects show significant variance depending on the provided perspective.}
    \label{tab:ablation_lmo}
\end{table}

Symmetric objects exhibit robustness to perspective changes, with a minimal average fluctuation ($\Delta$) of only 1.1 AP. In contrast, asymmetric objects are sensitive to the reference view; their performance variance reflects the difficulty of extracting a reliable identity prior when distinctive geometric features are occluded. Despite these instance level variations, the aggregate performance remains stable. As shown by the flat trend line in Fig.~\ref{fig:view_sensitive}, the mean average precision across the dataset fluctuates by a maximum of only 3.1 AP between the most and least favorable global perspectives.

\subsubsection{View Scaling.}

Previous experiments show that using only one reference view limits the performance on asymmetric objects. To address this, we leverage the high inference speed of our architecture to incorporate multiple reference views. Rather than introducing complex fusion modules, we perform independent inference for each available view to generate a mask prediction. For masks with significant spatial overlap (IoU $> 0.5$), we retain only the candidate with the highest confidence score. This strategy allows us to improve robustness without structural modifications.

\begin{figure}[!htbp]
    \centering
    \begin{minipage}[b]{0.45\textwidth}
        \centering
        \includegraphics[width=\textwidth]{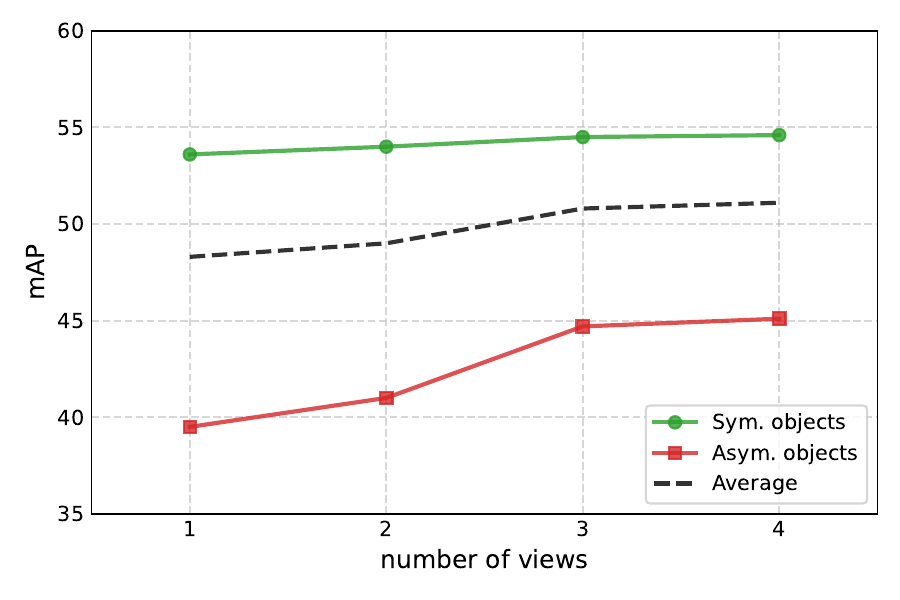}
        \centerline{\small (a) Internal scaling.}
    \end{minipage}
    \hfill
    \begin{minipage}[b]{0.45\textwidth}
        \centering
        \includegraphics[width=\textwidth]{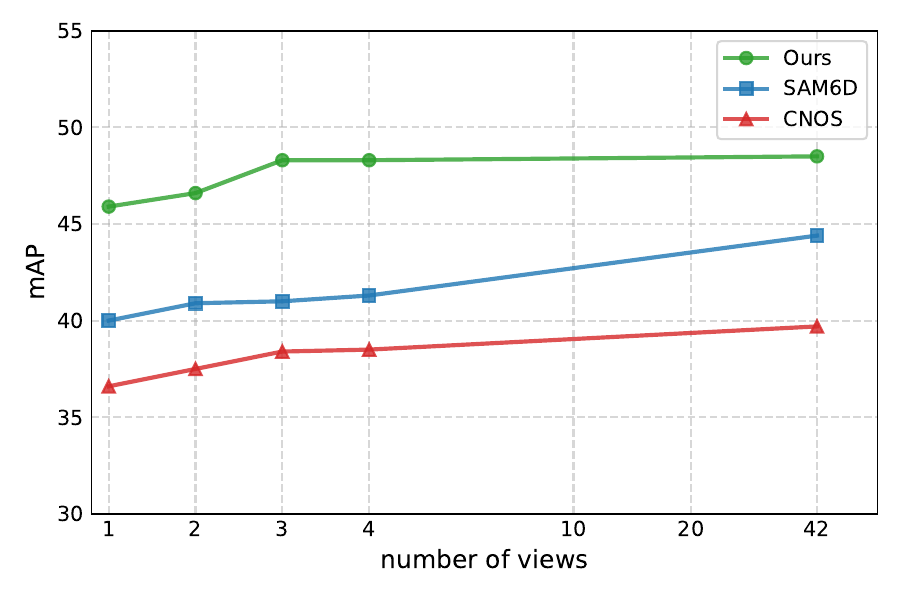}
        \centerline{\small (b) Comparative scaling.}
    \end{minipage}
    
    \caption{\textbf{Ablation on reference views.} (a) \textbf{Internal scaling}: Performance improves significantly by aggregating complementary geometric context, particularly for highly sensitive asymmetric objects, reaching a saturation point around three to four views. (b) \textbf{Comparative scaling}: Compared to baselines, our method achieves higher initial accuracy and reaches peak performance much earlier with only a fraction of the reference views.}
    \label{fig:ablation_studies}
\end{figure}

Fig.~\ref{fig:ablation_studies}(a) illustrates the quantitative results of this view scaling strategy. Most notably, this approach proves exceptionally effective for the asymmetric objects. By aggregating complementary geometric context, the average precision improves significantly when increasing the reference count from one to three views. The performance eventually reaches a saturation point around three to four views, after which additional images provide minimal accuracy gains.

We compare this scaling behavior against baselines in Fig.~\ref{fig:ablation_studies}(b). Our architecture establishes a higher initial accuracy with a single image and demonstrates a steeper improvement trajectory during early scaling. While baseline methods require dense templates to accumulate geometric context, our approach saturates much earlier with only a fraction of the views. This suggests that explicit multi template aggregation modules are computationally redundant when the underlying feature fusion is inherently robust.

\subsection{In-the-Wild Robustness}
\begin{figure}[!htbp]
    \centering
    \includegraphics[width=1.0\textwidth]{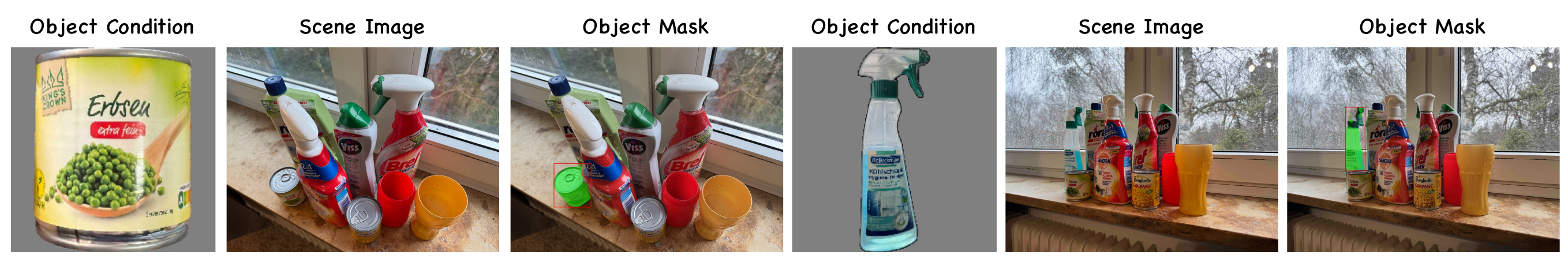}
    \caption{\textbf{Qualitative evaluation in unconstrained environments.} Our model accurately isolates target items in highly cluttered everyday scenes, handling severe occlusions and complex backgrounds entirely without domain specific training data.}
    \label{fig:in_the_wild}
\end{figure}

To demonstrate practical applicability, we evaluate \modelname in unconstrained, cluttered real-world environments. As shown in Fig.~\ref{fig:in_the_wild}, the model successfully segments everyday objects despite severe occlusions and significant viewpoint variations. 

Notably, \modelname demonstrates robust discriminative power against high inter-class similarity. In the left example, given a pea can as the object condition, our model accurately targets it while entirely bypassing the neighboring corn can and other distracting containers. Analogously, in the right example, when prompted with a specific detergent bottle, \modelname avoids false positives and successfully distinguishes the target from a row of multiple alternative cleaning products that possess highly similar shapes and labels. These results underscore the precise instance-level matching capability and robustness of our model in resolving real-world semantic ambiguities.

\section{Limitations} 
SOS currently focuses on the single-reference setting and assumes that each query image contains at most one target instance corresponding to the given reference. Therefore, it cannot handle cases where multiple identical target instances appear in the same query image. This limitation is partly due to the lack of suitable supervision. To the best of our knowledge, existing open-source reference-conditioned training datasets do not provide annotations for multiple identical target instances in one query image. Future work will extend SOS toward multi-instance reference-conditioned segmentation by introducing instance-aware prediction mechanisms and constructing training data with explicit multi-instance supervision.

\section{Conclusion}

In this work, we propose \textbf{\modelname}, a streamlined, model-free architecture for unseen objects segmentation. By conditioning on just one reference view, our method eliminates complex 3D rendering and achieves fast inference suitable for real-world applications. Despite this lightweight setup, our approach delivers accuracy comparable to state-of-the-art model-based methods, and crucially outperforms them by a significant improvement when they are restricted to the exact same single-view condition. Furthermore, empirical analysis demonstrates that without any domain-specific training, our model seamlessly accommodates sparse-view inputs to further elevate performance. Together, \modelname serves as a viable alternative to CAD-based designs for unseen objects segmentation which can pave the way to more broadly applicable practical setups.

\label{sec:conclusion}

\bibliography{egbib}

\end{document}